\documentclass[conference]{IEEEtran}
\IEEEoverridecommandlockouts
\usepackage{booktabs} 
\usepackage{threeparttable}
\usepackage{hyperref}
\usepackage{cite}
\usepackage{amsmath,amssymb,amsfonts}
\usepackage{algorithmic}
\usepackage{graphicx}
\usepackage{textcomp}
\usepackage{xcolor}
\usepackage{tikz}
\usepackage{float}
\usepackage{placeins}
\usetikzlibrary{shapes.geometric, arrows.meta, positioning, calc}
\usepackage{amsmath,amssymb}
\usepackage{graphicx}
\usepackage{booktabs}
\usepackage{xcolor}
\usepackage{float}
\usepackage{hyperref}
\usepackage{tikz}
\usetikzlibrary{arrows.meta,positioning,fit}

\definecolor{pblue}{HTML}{2A78D6}
\definecolor{porange}{HTML}{EB6834}
\definecolor{paqua}{HTML}{1BAF7A}
\definecolor{pprimary}{HTML}{0B0B0B}
\definecolor{psecondary}{HTML}{52514E}
\definecolor{pmuted}{HTML}{898781}

\tikzset{
  stagebox/.style={draw, thick, rounded corners=2pt, align=center,
    minimum height=1.7cm, inner sep=4pt, font=\footnotesize},
  chip/.style={draw=pmuted, rounded corners=1pt, align=center, fill=white,
    minimum height=1cm, inner sep=3pt, font=\scriptsize},
  groupbox/.style={draw, thick, rounded corners=2pt, inner sep=7pt},
  siglabel/.style={font=\scriptsize, text=psecondary, align=center},
}

\usepackage[utf8]{inputenc}
\usepackage{amssymb}
\usepackage{array,longtable}
\usepackage{booktabs}
\usepackage{enumitem}

\usepackage{pgfplots}
\pgfplotsset{compat=1.18}
\usepgfplotslibrary{groupplots}
\usepackage{subcaption}

\def\BibTeX{{\rm B\kern-.05em{\sc i\kern-.025em b}\kern-.08em
    T\kern-.1667em\lower.7ex\hbox{E}\kern-.125emX}}
\begin{document}

\title{Compliance for Free: Learning Identifiable Impedance via Bilateral Teleoperation\\
\thanks{Generative AI (Claude Code Sonnet 5 High) was used to generate code. Project website and code at https://icra2027.github.io/free-compliance/}
}


\author{
    \IEEEauthorblockN{Harsha Guda, Adri\`{a} Colom\'{e}, and Carme Torras}
    \IEEEauthorblockA{\textit{Institut de Robòtica i Informàtica Industrial, CSIC-UPC} \\
    Barcelona, Spain \\
    \{hguda, acolome, torras\}@iri.upc.edu}
}

\maketitle

\begin{abstract}
Vision-language-action models tell a robot where to move, but not how hard to push. Contact-rich tasks depend on that second quantity, compliance, yet no widely used demonstration interface records it. The obstacle is identifiability as realized pose and measured force cannot separate the operator's intended equilibrium from their stiffness, so VR controllers, SpaceMouse and handheld grippers cannot supply compliance supervision even in principle. Prior compliance-output policies work around this with hand-specified task structure, privileged simulation contact state, or dedicated force and tactile hardware. Four-channel bilateral teleoperation removes the ambiguity directly by using the leader arm as a separate measurement of the intended equilibrium, making per-axis stiffness identifiable by regression using only the joint-torque sensing already on the manipulator. This yields per-timestep, direction-dependent compliance labels at zero annotation cost, which we use to fine-tune a VLA to emit stiffness alongside pose. 
On a Franka Research 3 wiping task, ours is the only policy of five whose contact force changes when the instruction asks for a firm wipe rather than a normal one ($6.4$\,N (normal) $\to 9.1$\,N (firm) RMS, Cohen's $d = 0.89$, $p = 0.023$)
\end{abstract}

\section{INTRODUCTION}
Vision-Language-Action (VLA) models have revolutionized robot manipulation~\cite{brohan2022rt1, brohan2023rt2, ghosh2024octo, black2024pi0, shukor2025smolvla}, yet they consistently fail at the last millimeter. Contact-rich manipulation tasks such as wiping, tight tolerance insertion, and assembly require a robot to modulate how hard a robot pushes rather than just where it goes. In these scenarios, compliance is a first class action, every bit as critical as the target pose itself. 

The robotics community remains trapped in a data bottleneck regarding this physical interaction. Recent models have attempted to close the gap by injecting contact force as an additional input observation via a force-aware mixture of experts~\cite{yu2025forcevlaenhancingvlamodels} or hybrid control frameworks~\cite{li2026forcevla2}. However, the few architectures that actively output force or compliance targets still lean heavily on structured constraints. 
They require hand-specified interaction frames~\cite{fang2026force}, infer stiffness matrices from privileged simulation contact data~\cite{wang2026stiffnesscopilotimpedancepolicy}, or distill policies from dedicated inference-time tactile hardware~\cite{gubernatorov2026hapticvla}.

This barrier is not a failure of model architecture, but a fundamental mathematical flaw in the demonstration interfaces. The standard setups driving modern visuomotor policy learning such as low-cost bimanual rigs~\cite{zhao2023aloha}, universal manipulation interfaces~\cite{chi2024umi}, or VR controllers~\cite{khazatsky2024droid} record only the robot's realized pose. The consequence is easiest to understand as a spring. A soft spring pulled a long way and a stiff spring pulled slightly exert exactly the same force. Watching only where the robot ended up and how hard it pushed, one cannot tell which of the two the operator intended, unless something independently records where they were \emph{aiming}. Pose-only interfaces record no such signal, so the operator's intended pose and their stiffness remain confounded. This is a strict identifiability problem, these interfaces have never measured the impedance the human intended, and mathematically, they never will.

To resolve this ambiguity, we turn to four-channel bilateral control~\cite{lawrence1993stability}. Unlike passive recording interfaces, a bilateral setup actively couples a leader and a follower arm. We demonstrate that the leader-arm trajectory serves as an independent, physical measurement of the operator's intended equilibrium pose. By explicitly measuring what the human operator commands rather than just where the robot ends up, we collapse the equilibrium and stiffness ambiguity that has forced prior work into simulation. Critically, this resolution allows us to extract compliance without relying on dedicated force-torque or tactile sensors. Utilizing only the joint-torque sensing already present on the manipulator, we introduce a masked extraction procedure that yields per-timestep, anisotropic, real-world compliance labels. We subsequently use these labels to fine-tune a pretrained VLA to emit a stiffness target alongside each pose target. We validate on a whiteboard-wiping task instructed as ``wipe the \{left, right\} mark \{normally, firmly\}'', so each demonstration carries both a spatial target, which we call the \emph{referent}, and an instructed force character, which we call the \emph{manner}.

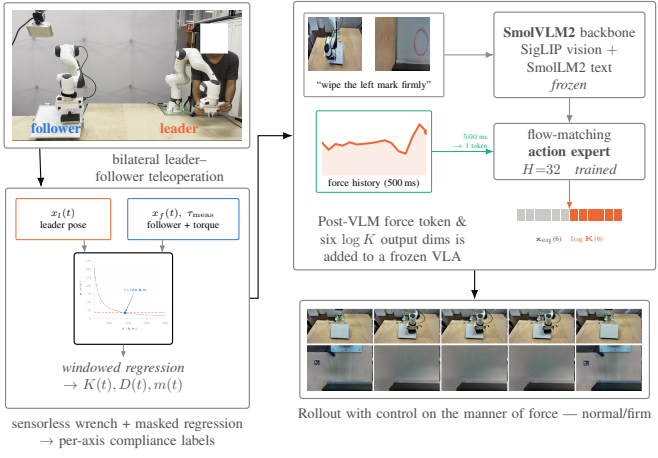
\begin{figure}[htbp]
  \centering
  \resizebox{\linewidth}{!}{%
\begin{tikzpicture}[node distance=6mm and 9mm]


\node (photo1) {\includegraphics[width=5.6cm]{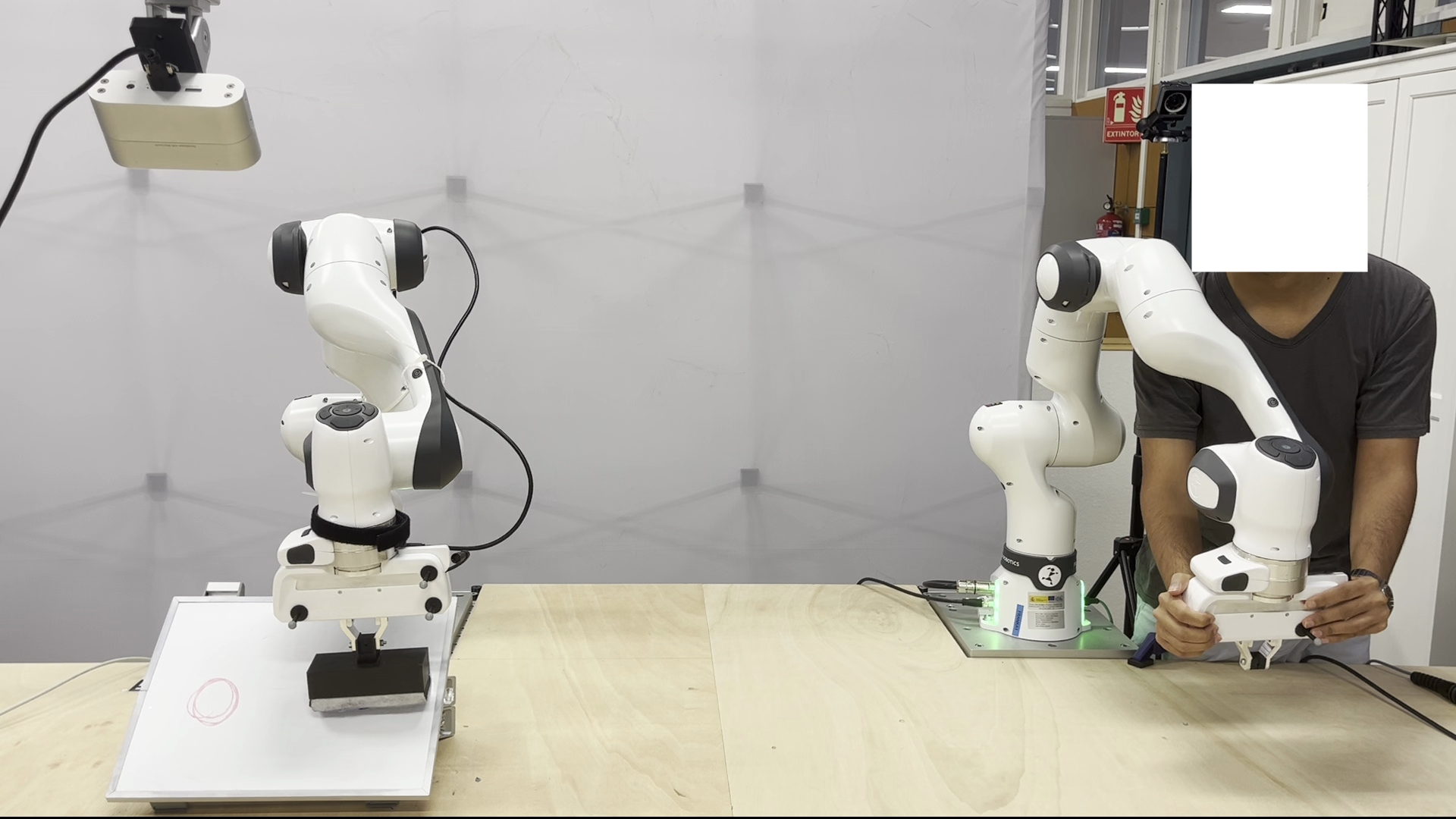}};
\node[porange, font=\normalsize\bfseries] (lbl-leader)
  at ($(photo1.south west)+(4.2cm,0.42cm)$) {leader};
\node[pblue, font=\normalsize\bfseries] (lbl-follower)
  at ($(photo1.south west)+(1.2cm,0.42cm)$) {follower};
\node[groupbox, draw=pmuted, fit=(photo1), inner sep=2pt,
  ] (collectbox) {};
\node[siglabel, below=0mm of collectbox, xshift=8mm, text width=3.2cm, font=\normalsize] (collectcap)
  {bilateral leader--follower teleoperation};


\node[chip, draw=porange,
  text width=2.1cm] (leaderpose) at (-15.5mm, -35mm) {$x_l(t)$\\leader pose};
\node[chip, right=4mm of leaderpose, draw=pblue, text width=2.5cm]
  (followerpose) {$x_f(t),\ \tau_{\mathrm{meas}}$\\follower + torque};
\coordinate (mid-le-fo) at ($(leaderpose.south)!0.5!(followerpose.south)$);
\node[stagebox, below=3mm of mid-le-fo, inner sep=3pt] (etbox)
  {\includegraphics[height=1.8cm]{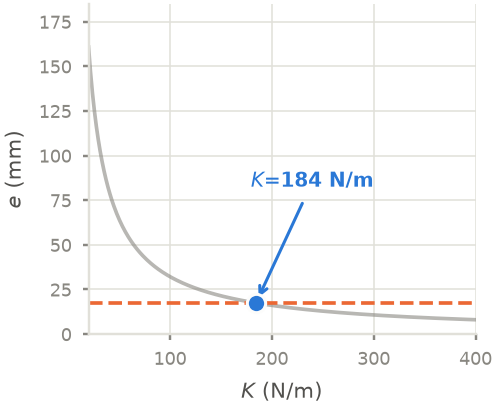}\\[-1mm]
   };
\draw[-{Latex[length=1.6mm]}, pmuted] (leaderpose.315) --  (etbox.130);
\draw[-{Latex[length=1.6mm]}, pmuted] (followerpose.225) -- (etbox.50);
\node[siglabel, below=3mm of etbox, text width=3.2cm, font=\normalsize\itshape]
  (windowed) {windowed regression $\rightarrow K(t), D(t), m(t)$};
\draw[-{Latex[length=1.4mm]}, pmuted] (etbox.south) -- (windowed.north);
\node[groupbox, draw=pmuted, fit=(leaderpose)(followerpose)(etbox)(windowed),
  ] (extractbox) {};
\node[siglabel, below=1mm of extractbox, text width=6cm, font=\normalsize]
  {sensorless wrench + masked regression $\rightarrow$ per-axis compliance labels};

\draw[-{Latex[length=2.4mm]}, thick, pprimary] (collectbox.220) -- (extractbox.128);


\node[chip, right=13mm of collectbox.east, yshift=5mm, draw=pmuted, inner sep=3pt]
  (scenecams) {%
    \begin{tabular}{@{}c@{\hspace{1mm}}c@{}}
      \includegraphics[width=1.3cm]{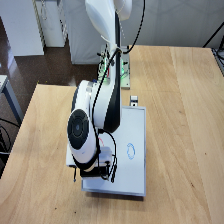} &
      \includegraphics[width=1.3cm]{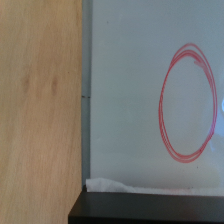} \\[0.5mm]
      \multicolumn{2}{c}{\scriptsize ``wipe the left mark firmly''}
    \end{tabular}%
  };
\node[chip, below=3.3mm of scenecams, draw=paqua, inner sep=3pt]
  (forcehist) {%
    \includegraphics[width=2.6cm]{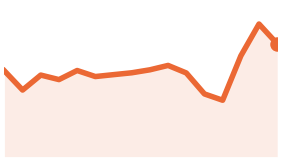}\\[-0.5mm]
    {\scriptsize force history (500\,ms)}%
  };

\node[stagebox, right=12mm of scenecams, draw=pmuted, text=psecondary,
      text width=3.4cm, minimum height=1.0cm, font=\normalsize] (vlmbox)
  {
      \\[0.4mm]
   \textbf{SmolVLM2} backbone\\SigLIP vision $+$ SmolLM2 text\\\emph{frozen}};

\node[stagebox, below=6mm of vlmbox, draw=pmuted, text=psecondary, text width=3.4cm,
      minimum height=1.0cm, font=\normalsize] (expertbox)
  {flow-matching \textbf{action expert}\\$H{=}32$\quad\emph{trained}};

\draw[-{Latex[length=1.8mm]}, pmuted] (scenecams.east) -- (vlmbox.west);
\draw[-{Latex[length=1.8mm]}, pmuted] (vlmbox.south)   -- (expertbox.north);
\draw[-{Latex[length=1.8mm]}, paqua]  (forcehist.east) -- ++(6mm,0) |- (expertbox.west);
\node[font=\tiny, text=paqua, anchor=south east, inner sep=1pt, align=right]
  at ($(expertbox.west)+(-1mm,0.4mm)$) {$500$\,ms\\$\rightarrow$ 1 token};

\node[below=7mm of expertbox, inner sep=0pt, align=center] (actvec) {%
  \textcolor{pmuted!45}{\rule{0.19cm}{0.28cm}}\hspace{0.6pt}%
  \textcolor{pmuted!45}{\rule{0.19cm}{0.28cm}}\hspace{0.6pt}%
  \textcolor{pmuted!45}{\rule{0.19cm}{0.28cm}}\hspace{0.6pt}%
  \textcolor{pmuted!45}{\rule{0.19cm}{0.28cm}}\hspace{0.6pt}%
  \textcolor{pmuted!45}{\rule{0.19cm}{0.28cm}}\hspace{0.6pt}%
  \textcolor{pmuted!45}{\rule{0.19cm}{0.28cm}}\hspace{1.4pt}%
  \textcolor{porange}{\rule{0.19cm}{0.28cm}}\hspace{0.6pt}%
  \textcolor{porange}{\rule{0.19cm}{0.28cm}}\hspace{0.6pt}%
  \textcolor{porange}{\rule{0.19cm}{0.28cm}}\hspace{0.6pt}%
  \textcolor{porange}{\rule{0.19cm}{0.28cm}}\hspace{0.6pt}%
  \textcolor{porange}{\rule{0.19cm}{0.28cm}}\hspace{0.6pt}%
  \textcolor{porange}{\rule{0.19cm}{0.28cm}}\hspace{1.4pt}%
  \\[0.7mm]
  {\tiny\textcolor{psecondary}{$\mathbf{x}_{\mathrm{eq}}$\,(6)}\hspace{2.2mm}%
   \textcolor{porange}{$\log\mathbf{K}$\,(6)}\hspace{2.2mm}%
   }\\[0.3mm]
};
\draw[-{Latex[length=1.6mm]}, porange] (expertbox.south) -- (actvec.north);

\node[groupbox, draw=pmuted,
  fit=(scenecams)(forcehist)(vlmbox)(expertbox)(actvec),
  ] (trainbox) {};
\node[siglabel, above=1mm of trainbox.south, xshift=-20mm, text width=3.8cm, font=\normalsize]
  {Post-VLM force token \& six $\log K$ output dims is added to a frozen VLA};

\draw[-{Latex[length=2.4mm]}, thick, pprimary] (extractbox.east) 
        -- ++(0,0) -| 
        ++(2mm, 0mm) |-
        (trainbox.west);


\node[below=8mm of trainbox.south] (photo4)
  {\includegraphics[height=2.0cm]{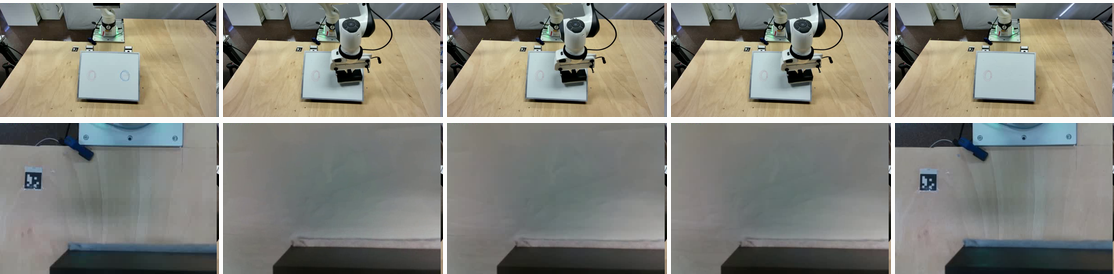}};
\node[groupbox, draw=pmuted, fit=(photo4), inner sep=2pt,
  ] (rolloutbox) {};
\node[siglabel, below=1mm of rolloutbox, text width=9cm, font=\normalsize]
  {Rollout with control on the manner of force --- normal/firm};

\draw[-{Latex[length=2.4mm]}, thick, pprimary] (trainbox.south) -- (rolloutbox.north);



\end{tikzpicture}
} 
  \caption{From bilateral demonstrations to instruction-conditioned compliance.}
  \label{fig:teaser}
\end{figure}

In summary, this paper makes the following contributions:
\begin{itemize}
    \item A sensorless extraction procedure that resolves the pose-only identifiability ambiguity via bilateral teleoperation, yielding per-timestep compliance labels using only native joint-torque sensors.
    \item An offline benchmark showing that the labels carry learnable signal: a trained predictor beats a per-axis constant predictor on every translational axis, for a $1.25\times$ aggregate improvement.
    \item A compliance-output VLA policy trained on these extracted labels that safely tracks chunked impedance targets and is the only policy of five whose realized
    contact force responds to the manner instruction.
\end{itemize}

\section{RELATED WORK}
\label{sec:related_work}
\subsection{Force-aware and compliance-output policies}

A large body of vision-language-action (VLA) policies such as RT-1~\cite{brohan2022rt1}, RT-2~\cite{brohan2023rt2}, Octo~\cite{ghosh2024octo}, $\pi_0$~\cite{black2024pi0}, and SmolVLA~\cite{shukor2025smolvla} learn a generalist policy mapping from vision and language to robot actions at scale. However, each of them only outputs pose, joint, or gripper targets. The stiffness for the low-level controller is fixed, not usually predicted by the policy.  

Recent attempts to close this gap have moved beyond fixed stiffness. Models like ForceVLA~\cite{yu2025forcevlaenhancingvlamodels} and FD-VLA~\cite{zhao2026fd} incorporate contact force as an input feature, either via mixture-of-experts or cross-modal distillation, to make position outputs 'force-aware'. Meanwhile, hybrid frameworks like ForceVLA2~\cite{li2026forcevla2} predict target forces alongside position. However, these methods either treat force as a rigid target rather than variable compliance, or require a dedicated force/torque (F/T) sensor at the wrist during both training and inference.

Among policies that produce compliance without F/T hardware, each leans on structure this paper's setting lacks. Force Policy \cite{fang2026force} relies on a hand-specified interaction frame. Stiffness Copilot \cite{wang2026stiffnesscopilotimpedancepolicy} infers direction-dependent stiffness matrices in simulation using privileged contact information. And recent approaches like PaCo-VLA \cite{cao2026paco} structure compliance via hand-designed task stages and semantic bindings. Although, HapticVLA~\cite{gubernatorov2026hapticvla} removes tactile sensing at inference by distilling from real tactile supervision, it still needs a tactile sensor at time of training. Our method bypasses these constraints by extracting pure, unrestricted impedance targets directly from human demonstrations.

Bi-LAT~\cite{kobayashi2025bi} is closest in spirit, it also learns from bilateral-control demonstrations and modulates applied force by language (``softly grasp'' vs.\ ``strongly twist''), but it regresses joint torque directly rather than extracting an explicit, physically interpretable stiffness label, so its notion of force control does not rest on an identifiability argument and cannot be reused to supervise a different backbone or task. None of these five requires zero dedicated force/tactile hardware at both training and inference time, and none derives compliance supervision from an identifiability argument applicable to any existing bilateral-teleoperation dataset, the gap this paper's extraction procedure (Section~\ref{sec:identifiability}) and policy (Section~\ref{sec:policy}) fill.
Closest to our setting, Xu et al.~\cite{xu2026mindthegap} also read impedance out of a bilateral leader--follower rig without force sensing, treating the mismatch between leader and follower response as an implicit encoding of interaction force, and cloning operator \emph{intent} rather than realized motion. Their impedance remains implicit, no stiffness label is ever produced.

Each of these workarounds substitutes for something the demonstrations never carried. The interfaces producing most VLA training data instrument only the robot's realized trajectory, so the structure these methods import either from simulation, from a hand-specified frame, or from extra hardware, as a stand in for an equilibrium signal absent at collection time. Leader--follower rigs are not automatically exempt: ALOHA's arms~\cite{zhao2023aloha} are unilateral, rendering no contact force back to the operator, so the leader--follower deflection reflects the servo's tracking error under its own gains rather than an operator yielding to felt contact. Bilateral coupling is what turns that deflection into a measurement.

\subsection{Bilateral Teleoperation and Control}
In a four-channel bilateral architecture, leader and follower exchange both position and force in both the directions: the follower tracks the operator's hand, and the contact the follower feels is rendered back at the operator. This is a mature control technique, with stability and transparency thoroughly analyzed since Lawrence~\cite{lawrence1993stability}. We leverage this theoretical maturity rather than developing bilateral control itself as a core contribution. 

Recent work has begun exploring bilateral setups for imitation learning data collection. Yamane et al.~\cite{yamane2026design} build a sensorless 4-channel framework using disturbance observers to estimate force for manipulators lacking joint-torque sensors, while Inami et al.~\cite{inami2025motion} exploit position/force duality to retroactively edit recorded motions. Prometheus~\cite{satsevich2025prometheus} introduces a mocap-based teleoperation system, but requires adding a dedicated compression force sensor at the gripper for operator feedback.

Our approach diverges from this recent literature in two fundamental ways:
\begin{itemize}
    \item Unlike Prometheus~\cite{satsevich2025prometheus}, our wrench estimation and compliance labels utilize only the manipulator's native joint-torque sensors, requiring zero additional force/torque hardware.
    \item Unlike Yamane et al.~\cite{yamane2026design}, we do not merely estimate a scalar external force. We extract an explicit, interpretable per-axis stiffness matrix to supervise a compliance-output VLA policy.
\end{itemize}

\section{Methodology}
\label{sec:methodology}

\subsection{The Identifiability Problem}
\label{sec:setup}

An impedance law says the arm behaves like a spring and damper: the further it is displaced from where it is trying to be, the harder it pushes back, and the faster it is displaced, the more it resists. We model contact at the end-effector with a diagonal, time-varying Cartesian impedance law
\begin{equation}
\label{eq:impedance-law}
\mathbf{f}(t) = \mathbf{K}(t)\,\mathbf{e}(t) + \mathbf{D}(t)\,\dot{\mathbf{e}}(t),
\qquad
\mathbf{e}(t) = \mathbf{x}_{\mathrm{eq}}(t) - \mathbf{x}_f(t),
\end{equation}
where $\mathbf{x}_f(t)$ is the measured follower pose and $\mathbf{f}(t)$ the measured contact wrench, the six-axis vector of forces and torques at the end-effector. $\mathbf{K}(t) = \operatorname{diag}(k_1,\dots,k_6)$ and $\mathbf{D}(t) = \operatorname{diag}(d_1,\dots,d_6)$ are stiffness and damping in the contact frame. The remaining term $\mathbf{x}_{\mathrm{eq}}(t)$ is the operator's commanded equilibrium pose, where the arm is \emph{trying} to be
rather than where it actually is, and it is the one quantity in Eq.~\eqref{eq:impedance-law} that a pose-only interface never records.

That omission is a strict identifiability failure, not a data shortage. Take a single axis $i$ and drop damping for a moment, so $f_i = k_i e_i$. For any scale $\alpha > 0$ the pair $(\alpha k_i,\, e_i/\alpha)$ reproduces the observed $f_i$ exactly, and since $x_f^i$ is measured, adopting the displacement $e_i/\alpha$ just amounts to asserting the operator aimed at $x_{\mathrm{eq}}^i = x_f^i + e_i/\alpha$. Every positive stiffness is therefore consistent with the recording under \emph{some} intended equilibrium. 
For a pose-only interface, then, the demonstrator's intended equilibrium and the robot's realized pose are the same recorded number, so no volume of such data recovers $\mathbf{K}$ after the fact. Doing so requires a separate measurement of $\mathbf{x}_{\mathrm{eq}}$.

\subsection{Resolving Identifiability via Bilateral Teleoperation}
\label{sec:identifiability}
Four-channel bilateral teleoperation fundamentally alters this paradigm. By actively coupling a leader and follower arm, the system records a second, separately instrumented pose stream that the follower's own state does not determine, and we take that stream as the operator's commanded equilibrium: $\mathbf{x}_{\mathrm{eq}}(t) := \mathbf{x}_l(t)$, the leader pose. Substituting it collapses the ambiguity, $\mathbf{e}(t) = \mathbf{x}_l(t) - \mathbf{x}_f(t)$ is now directly observed, and $(\mathbf{K}(t), \mathbf{D}(t))$ become identifiable by regression on $(\mathbf{e}, \mathbf{\dot{e}}, \mathbf{f})$. In other terms, we can now see both how far the spring was pulled and how hard it pulled back, which fixes its stiffness. This holds only where the data is in the direction of the axis. If the arm barely deflects, or barely pushes, the regression has nothing to work with. The identifiability mask defined below makes that requirement explicit, per axis and per timestep.

\subsection{Compliance label extraction}
\subsubsection*{Contact-frame definition}
Stiffness is only meaningful relative to a frame, so we first recover the
geometry of the contact. Wiping strokes sweep out a plane. The one direction
they never explore is the direction pointing out of the board. The smallest
singular vector of the contact positions therefore recovers the surface
normal, and the ratio of smallest to largest singular value measures how
plane-like the motion really was.

We fit a contact frame for each demonstration, or pooled across a session when the physical contact surface is fixed throughout. Given a set of in-contact follower positions $\mathbf{p}_j \in \mathbb{R}^3$ with mean $\bar{\mathbf{p}}$, we take the singular value decomposition of the centered positions $\{\mathbf{p}_j - \bar{\mathbf{p}}\}$ and set $\hat{\mathbf{n}} = \mathbf{v}_3$, the right singular vector associated with the smallest singular value $\sigma_3$, with planarity $= \sigma_3/\sigma_1$. A fit is rejected when planarity $> 0.15$, i.e.\ when the motion was not planar enough to define a surface.

\textbf{Windowed regression for $(k, d)$:} For each short slice of the
demonstration we ask which spring-and-damper pair best explains the force
measured over that slice. Concretely, for each axis $i$, we estimate $(k_{i}, d_{i})$ over a sliding 300ms window directly at the 1kHz control rate by solving a regularized, box-constrained nonlinear least-squares problem:
\begin{equation}
\label{eq:windowed-regression}
\min_{k_{i}, d_{i}} \sum_{t \in W} \left( f_{i}(t) - k_{i}e_{i}(t) - d_{i}\dot{e}_{i}(t) \right)^{2} + \lambda \| \log k_{i} - \log k_{i}^{prior} \|^{2}
\end{equation}
subject to $k_{i} \in [k_{min}, k_{max}]$ and $d_{i} \ge 0$. Because the regularizer is nonlinear in $k_{i}$, this does not admit a closed-form ridge solution and is solved in log-space. The bounds $[k_{min}, k_{max}]$ are strictly defined by the realistic safety limits of the physical Cartesian impedance controller.
We use critically damped $\mathbf{D}$ values derived from $\mathbf{K}$ (${d_i} = 2\sqrt{{k_i}}$).

\subsubsection*{Identifiability mask} 
\begin{figure*}[t]
\centering
\begin{subcaptionbox}{Extracted stiffness traces $\mathbf{K}(t)$ throughout in-contact phase.\label{fig:stiffness-traces}}[0.32\linewidth]
{
  \includegraphics[width=\linewidth]{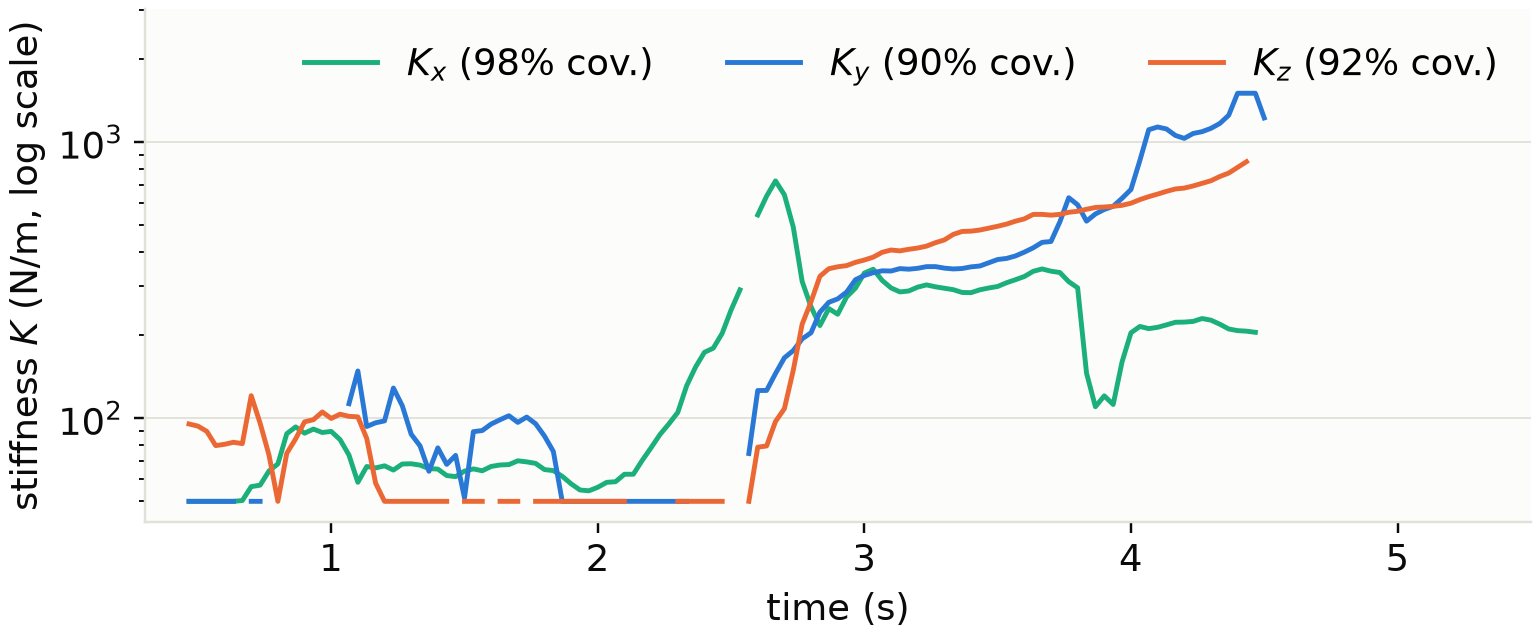}
}
\end{subcaptionbox}
\hfill
\begin{subcaptionbox}{Identifiability mask coverage, per axis, dataset-wide, broken out by manner.\label{fig:mask_coverage}}[0.32\linewidth]
{\includegraphics[width=\linewidth]{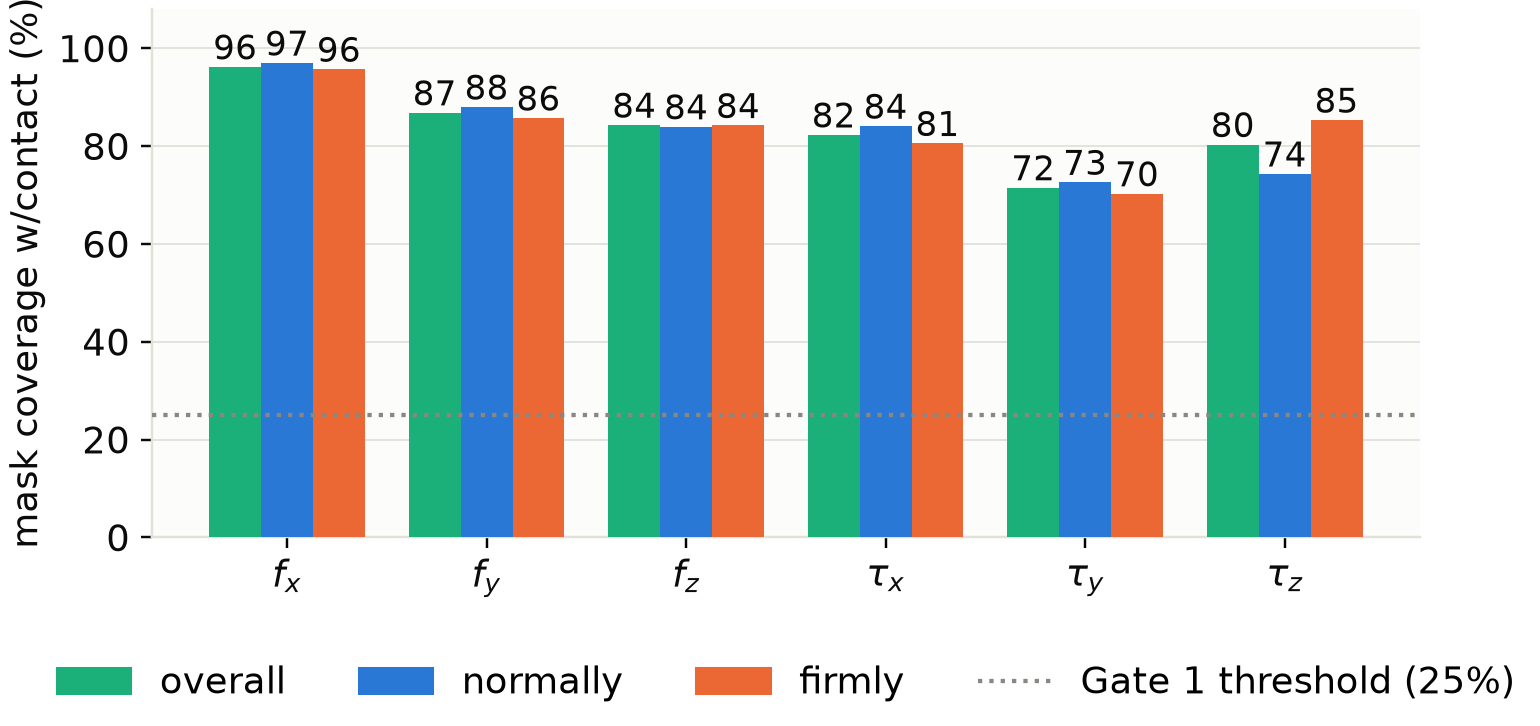}}
\end{subcaptionbox}
\hfill
\begin{subcaptionbox}{Sensorless wrench noise floor $\sigma_{f,i}$
  (Eq.~\eqref{eq:noise-floor}), measured from free-space residuals on
  held-out sweep sessions.\label{fig:noise-floor}}[0.32\linewidth]
{\includegraphics[width=\linewidth]{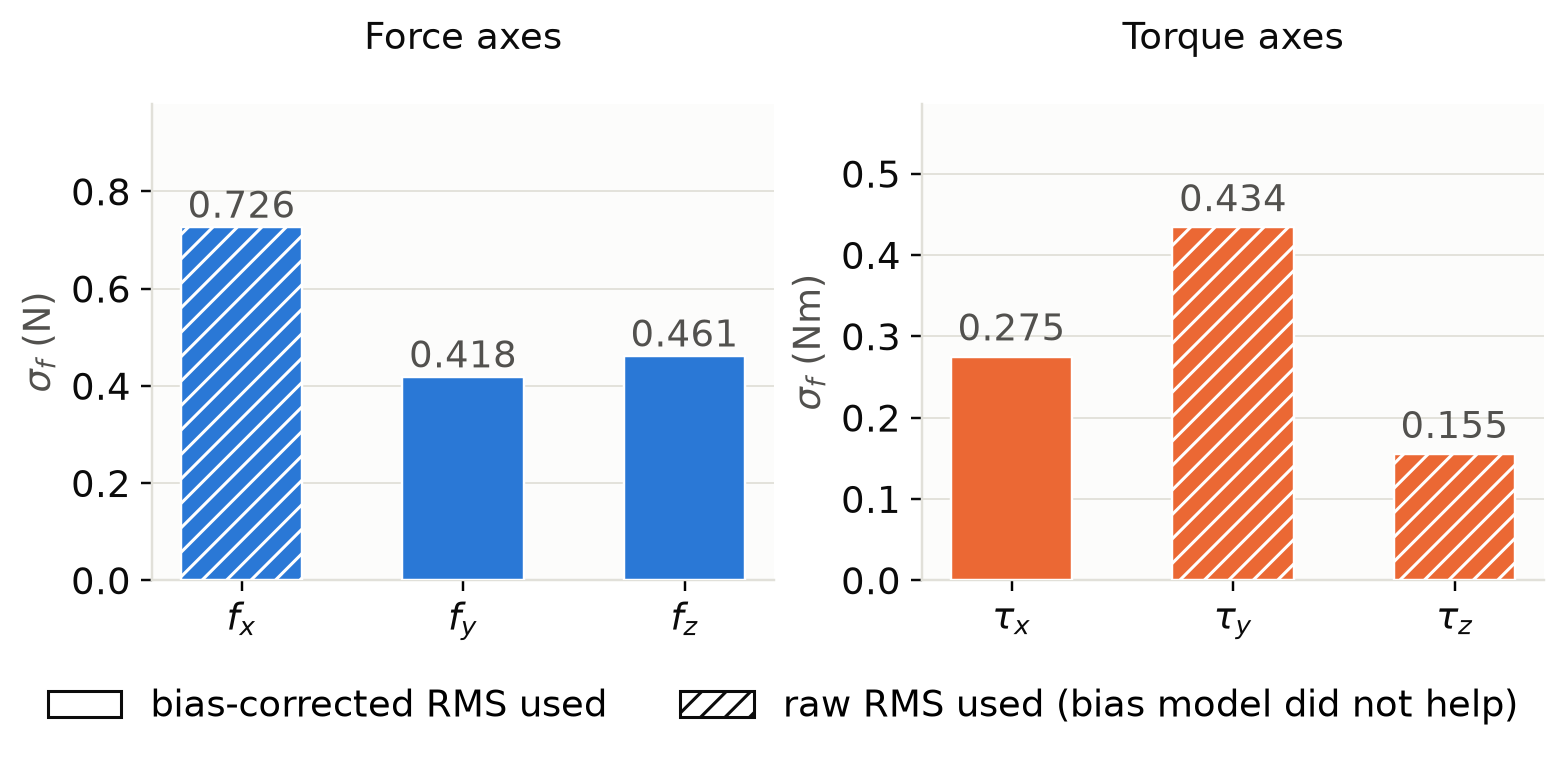}}
\end{subcaptionbox}
\caption{Dataset validity, dataset-wide. Together these establish that the
extracted compliance labels are identifiable with real margin (a), retained
at high coverage after masking (b), and
measured against a real, held-out noise floor well below the labels'
working range rather than assumed or hand-tuned (c). Hatched
bars mark axes where the bias-corrected estimate did not improve over the
raw RMS ($f_x$, $\tau_y$, $\tau_z$), for which $\sigma_{f,i}$ is the raw
value; this is a free-space noise floor.}
\label{fig:dataset-validity}
\end{figure*}

A regression will always return \emph{some} answer, even where the data cannot support one. We therefore keep a label only where it could genuinely have been revealed that the arm must be in contact, deflected far enough for the displacement to mean something, pushing harder than the sensor's noise, and in a configuration where the regression is numerically well conditioned.
For each axis $i$ and timestep $t$, a binary validity mask is defined as follows,

\begin{equation}
\label{eq:mask}
m_i(t) =
\begin{cases}
1 & \text{if } \kappa(\mathbf{J}_W^\top\mathbf{J}_W) < \kappa_{\max},\; |e_i(t)| > \sigma_e, \\
  & \phantom{\text{if }} |f_i(t)| > \sigma_{f,i},\;
      \text{and } \mathrm{contact}(t), \\
0 & \text{otherwise.}
\end{cases}
\end{equation}
combining, in that order, a regression-conditioning term ($\kappa$ is the condition number of the window's regressor matrix, large when the two columns are nearly collinear and the fit is unstable), a minimum-displacement term, a per-axis force-above-noise-floor term (from $\sigma_{f,i}$ by Eq.~\eqref{eq:noise-floor} ), and a contact indicator. Figure~\ref{fig:mask_coverage} shows per-axis mask coverage, computed within in-contact timesteps excluding free-space and reset time motion. 

\subsection{Wrench Estimation}
\label{sec:wrench}
Everything above assumes we can measure the contact wrench $\mathbf{f}$. We obtain it without a force sensor, from a simple observation. If a joint needs more torque than gravity alone would account for, something external must be pushing on the tool. Turning that into a calibrated wrench takes two steps --- identifying the tool's own weight, so it is not mistaken for contact, and then modelling what residual error remains. At static poses, $(\dot{\mathbf{q}} = \ddot{\mathbf{q}} = \mathbf{0})$, the residual between measured joint torque and the arm's zero-payload gravity model is linear in the unknown payload's mass and mass-weighted center of mass:

\begin{equation}
\label{eq:payload-id}
\Delta\boldsymbol{\tau}(\mathbf{q}) = \boldsymbol{\tau}_{\mathrm{meas}}(\mathbf{q}) - \boldsymbol{\tau}_{\mathrm{grav},0}(\mathbf{q})
= \mathbf{A}(\mathbf{q}) \begin{bmatrix} m \\ m c_x \\ m c_y \\ m c_z \end{bmatrix},
\end{equation}
where $\mathbf{A}(\mathbf{q}) \in \mathbb{R}^{7\times 4}$ is a gravity regressor. We solve it by least squares over $15$ static poses, in the sign convention in which a positive $m$ is an added mass pulling down. This is the same identifiability logic as Sections~\ref{sec:setup},~\ref{sec:identifiability}.

Immediately before every session we sample the wrench bias at a fixed free-space pose and subtract it as a per-axis constant offset for the remainder of the session. What remains after that offset is a slowly varying error that depends on where the arm is and how fast it is moving. We learn it from free-space sweeps, where by construction there is no contact and any reported force is error. Projecting the input through a bank of $D_f$ random Fourier features $\boldsymbol{\varphi}(\mathbf{x})$~\cite{rahimi2007random} makes a flexible nonlinear fit available in closed form, at a fraction of the cost of the Gaussian-process regression it approximates:
\begin{equation}
\label{eq:rff-ridge}
\mathbf{f}_{\mathrm{bias}}(\mathbf{x}) \approx \mathbf{W} \boldsymbol{\varphi}(\mathbf{x}), \qquad
\mathbf{W} = \operatorname*{argmin}_{\mathbf{W}} \lVert \mathbf{F} - \mathbf{W}\boldsymbol{\Phi} \rVert^2 + \lambda \lVert \mathbf{W} \rVert^2,
\end{equation}
solved in closed form and fit independently per axis on $\mathbf{x} = (\mathbf{q}, \dot{\mathbf{q}}) \in \mathbb{R}^{14}$, with $D_f = 512$ features, length scale $\ell = 30$ and ridge penalty $\lambda = 0.1$.

Because estimation error is strongly configuration-dependent, we pin a fixed nullspace --- a seven-joint arm can hold the same tool pose with many different elbow configurations, and we always choose the same one --- posture and constrain the working volume during both calibration and demonstration collection.

We define the per-axis noise floor, evaluated on held-out free-space sweep sessions, as
\begin{equation}
\label{eq:noise-floor}
\sigma_{f,i} = \min\!\left(\mathrm{RMS}_i^{\mathrm{raw}},\; \mathrm{RMS}_i^{\mathrm{bias\text{-}corrected}}\right),
\end{equation}
i.e.\ the bias-corrected estimate is used only on axes where it measurably improves over the raw estimate. $\sigma_{f,i}$ is used by mask condition~(iii) in Eq.~\eqref{eq:mask}.

\subsection{Label-quality validation: Offline compliance prediction benchmark}
This benchmark tests whether the extracted labels carry any meaningful signal. The task is to predict $\log K_i(t)$, at masked identifiable timesteps only, from a $20$-dimensional feature vector $\mathbf{x}_t = [\,\mathbf{q}_t,\, \dot{\mathbf{q}}_t,\, \text{manner},\, \text{referent}\,]$ that deliberately excludes force, so that the check cannot be circular. We compare two predictors: a \emph{constant} baseline, the per-axis training-split mean of $\log k_i$, and a \emph{learned} one, the random-Fourier-feature ridge model of Eq.~\eqref{eq:rff-ridge} reused here for a second, distinct regression target. We report per-axis RMSE on $\log \mathbf{K}$ over the masked test set, aggregated by averaging across axes.

Taken together, Sections~\ref{sec:identifiability}--\ref{sec:wrench} turn one recorded bilateral demonstration into a pair of arrays at the policy's own $30$\,Hz action rate: $\log \mathbf{K} \in \mathbb{R}^{N\times 6}$, expressed in that episode's fitted contact frame, and a Boolean identifiability mask of the same shape marking which of those entries the data actually supports. Each row is indexed to the raw frame nearest its window, so every label lines up with the image and proprioception the policy conditions on.

\subsection{Compliance-output Policy}
\label{sec:policy}

\textbf{Backbone.} SmolVLA~\cite{shukor2025smolvla} ($\sim 450$M parameters from LeRobot), with the pretrained vision-language model frozen and the action expert fine-tuned.

\textbf{Inputs.} We utilize two $224 \times 224 \times 3$ RGB streams: a scene camera capturing the global workspace and a rigidly mounted wrist camera. A language instruction of ``wipe the \{referent\} mark \{manner\}'', where referent $\in$ \{left, right\} and manner $\in$ \{normally, firmly\}. We identify the target by position (``left''/``right'') rather than by color, because the red and blue marks in our demonstrations occupy very few pixels and the model could not reliably tell them apart. A force-history token, a trailing 500ms of 6-DoF wrench downsampled to 20 samples (encoded by a 1D-convolutional network) is injected post-VLM (following the ForceVLA~\cite{yu2025forcevlaenhancingvlamodels}).

\textbf{Output.} The policy outputs an \emph{action chunk} of $H=32$ future timesteps predicted in one forward pass, rather than a single next step, each carrying $[\mathbf{x}_{\mathrm{eq}}(6),\, \log \mathbf{K}(6)]$. $\mathbf{x}_{\mathrm{eq}}(6)$ is an absolute target pose in the same base frame convention as $\mathbf{x}_f$. We predict absolute rather than relative poses because the identifiability argument of this paper depends on $\mathbf{x}_{\mathrm{eq}}$ being directly comparable to an absolute pose.

\textbf{Loss.} Flow-matching (L1) on $\mathbf{x}_{\mathrm{eq}}$, and on $\log \mathbf{K}$ a Huber loss~\cite{huber1992robust} --- quadratic near zero and linear in the tails, so occasional badly fit windows do not dominate the gradient --- applied only at timesteps the mask marks identifiable, at a weight tuned on the validation split.

\textbf{Augmentation.} Force dropout ($p = 0.15$) and random wrench-bias injection ($\pm 1$\,N) during training, to guard against modal masking, a failure mode in which the high-dimensional visual stream dominates optimization and low-dimensional force is silently ignored despite being available during training.

\textbf{Training.} All policies are fine-tuned identically: $2500$ steps, batch size $64$, AdamW at a constant $10^{-4}$ ($\beta = (0.9, 0.95)$, weight decay $10^{-10}$, gradient-norm clip $10$), three seeds, $99.9$M trainable parameters of $450.1$M. Proprioception is $[\mathbf{q}\,(7), \dot{\mathbf{q}}\,(7), \mathbf{x}_f\,(6)]$. Trained on a single NVIDIA L40S.

\subsection{Low-level controller}
\label{sec:controller}
The Franka Research 3 follower arm is driven at $1$\,kHz by a Cartesian impedance controller, built on the CRISP ROS\,2 compliant-controller stack~\cite{pro2026crisp},

\begin{equation}
\label{eq:controller-law}
\boldsymbol{\tau} = \mathbf{J}^\top\!\big[\mathbf{K}_t(\mathbf{x}_{\mathrm{eq}} - \mathbf{x}) + \mathbf{D}_t(\dot{\mathbf{x}}_{\mathrm{eq}} - \dot{\mathbf{x}})\big]
+ \mathbf{C}(\mathbf{q}, \dot{\mathbf{q}})\dot{\mathbf{q}} + \boldsymbol{\tau}_{\mathrm{null}},
\end{equation}

The controller tracks the policy's chunked $(\mathbf{x}_{\mathrm{eq}}(t),\, \mathbf{K}(t))$ output at 1\,kHz while the policy replans at $\sim$10\, Hz (taking 3 actions per chunk to execute).

\textbf{Temporal ensembling.} Successive chunks are inferred independently, so nothing forces the end of one to line up with the start of the next. Discarding the previous chunk each time a new one arrives creating a discontinuity that a rigid, non-compliant policy cannot physically absorb, and which stalls trajectory execution. We use ACT's temporal ensembling~\cite{zhao2023aloha} unchanged, combining every buffered chunk's prediction for a given low-level step by exponential recency weighting,
\begin{equation}
\label{eq:temporal-ensemble}
\hat{\mathbf{a}}(t) = \frac{\sum_i w_i\, \mathbf{a}_i(t)}{\sum_i w_i}, \qquad
w_i = \exp(-m \cdot \mathrm{age}_i(t)),
\end{equation}
where $\mathbf{a}_i(t)$ is chunk $i$'s prediction for step $t$ and $\mathrm{age}_i(t)$ is the number of low-level steps since chunk $i$ was queried, so the freshest available prediction for a given step dominates, while older overlapping predictions still smooth the boundary rather than being discarded outright. Position, orientation (as a rotation vector), and $\log \mathbf{K}$ are all blended this way. $m$ is tuned empirically, not assumed correct a priori.

\textbf{Stiffness rate limiting.} Piecewise-constant $\mathbf{K}$ arriving once per chunk injects energy at the boundary and can make the arm buzz or trip a protective stop, so we rate-limit in log-space, $|\mathrm{d}\log\mathbf{K}/\mathrm{d}t| \leq \gamma$, with a final tuned $\gamma = 3\,\mathrm{s}^{-1}$ on all six axes.

\textbf{Passivity.} Raising $\mathbf{K}$ injects energy, so stiffness
\emph{increases} are gated by a standard energy tank~\cite{ferraguti2013tank,schindlbeck2015unified}
($E_0 = 1$\,J, $E_{\max} = 5$\,J); decreases are always permitted.

\section{EXPERIMENTS AND RESULTS}
\label{sec:results}
\begin{figure*}[t]
    \centering
    \includegraphics[width=0.8\linewidth]{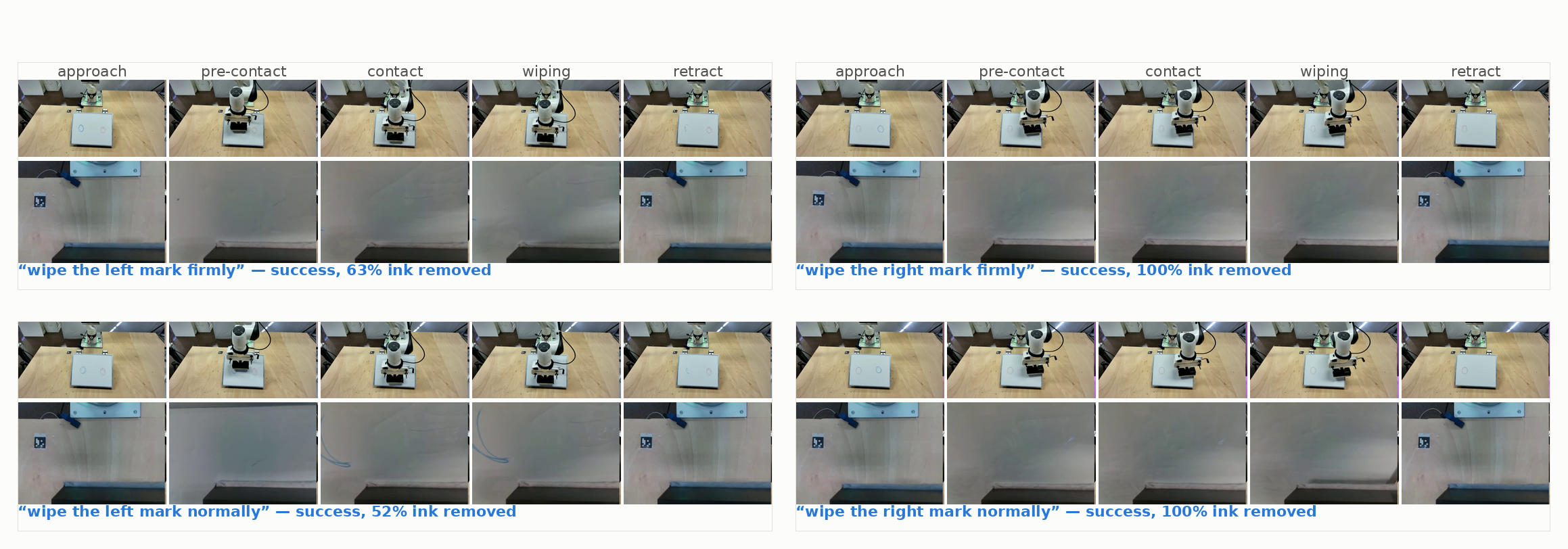}
    \caption{Qualitative rollouts of our policy, scene camera (top sub-row) and wrist camera (bottom sub-row), across the 2$\times$2 referent$\times$manner instruction grid (``wipe the left/right mark firmly/normally''), five phases per rollout (approach, pre-contact, contact, wiping, retract). Each cell shows one representative rollout ($n=1$). Aggregate success rates across all rollouts are reported in Table~\ref{tab:e1-primary}.}
    \label{fig:qualitative_rollouts}
\end{figure*}
\subsection{Experimental Setup}
\label{sec:setup-exp}

All policies are evaluated on a whiteboard wiping task while the board is at ${\sim}45^\circ$ from vertical, with a rigid eraser mounted to the end-effector of the manipulator, over a $2\times2$ instruction grid ``wipe the \{left, right\} mark \{normally, firmly\}'', $n=8$ rollouts per cell, three training seeds, $24$ rollouts per policy. Every rollout runs to a fixed $200$ low-level steps (${\approx}43.5$\,s); the policy replans every three executed chunk steps at a $30$\,Hz action rate (${\approx}10$\,Hz replanning), with temporal ensembling (Eq.~\eqref{eq:temporal-ensemble}, $m=0.05$) enabled identically for all policies. We report three quantities. \textbf{Success} is ink removal $\geq 50\%$, scored automatically by pixel difference from a fixed scene camera, so wiping the wrong mark is penalized rather than credited. \textbf{Ink removal} is that same percentage, continuous. \textbf{Contact force} is peak and RMS from the estimated wrench. \textbf{Prot. Stops} is a \emph{proxy} for a protective stop when peak estimated contact force is above a $40$\,N safety threshold. Table~\ref{tab:policies} lists the five policies compared.

\begin{table}[t]
\centering
\caption{Policies compared. All five share demonstrations, low-level
controller, seeds and replanning settings. They differ only in the inputs and outputs.}
\label{tab:policies}
\begin{threeparttable}
\footnotesize
\begin{tabular}{llll}
\toprule
Policy & Force in & Action output & Basis \\
\midrule
Fixed-K       & ---        & pose            & hand-set $\mathbf{K}$\tnote{a} \\
Fitted-K      & ---        & pose            & $\mathbf{K}$ fit to our labels\tnote{b} \\
Force-In      & \checkmark & pose            & ForceVLA~\cite{yu2025forcevlaenhancingvlamodels} \\
Hybrid        & \checkmark & pose $+ \log \mathbf{K}$\tnote{c} & ForceVLA2~\cite{li2026forcevla2,fang2026force} \\
\textbf{Ours} & \checkmark & pose $+ \log \mathbf{K}$\tnote{d} & extracted labels \\
\bottomrule
\end{tabular}
\begin{tablenotes}\footnotesize
\item[a] $(300,300,300)$\,N/m, $(20,20,20)$\,N$\cdot$m/rad --- a conventional,
well-posed constant impedance.
\item[b] Per-axis constants best fitting our label distribution
(Sec.~\ref{sec:results-oracle}).
\item[c] Pose decoded by flow matching, $\log \mathbf{K}$ by a \emph{separate}
one-shot linear regression head on the same suffix representation --- the
generative-pose / direct-regression-control split these systems use.
\item[d] Pose and $\log \mathbf{K}$ decoded through a single shared
flow-matching target, split only at the loss. This is what separates Ours from Hybrid.
\end{tablenotes}
\end{threeparttable}
\end{table}

\begin{figure}[t]
    \centering
    \includegraphics[width=\linewidth]{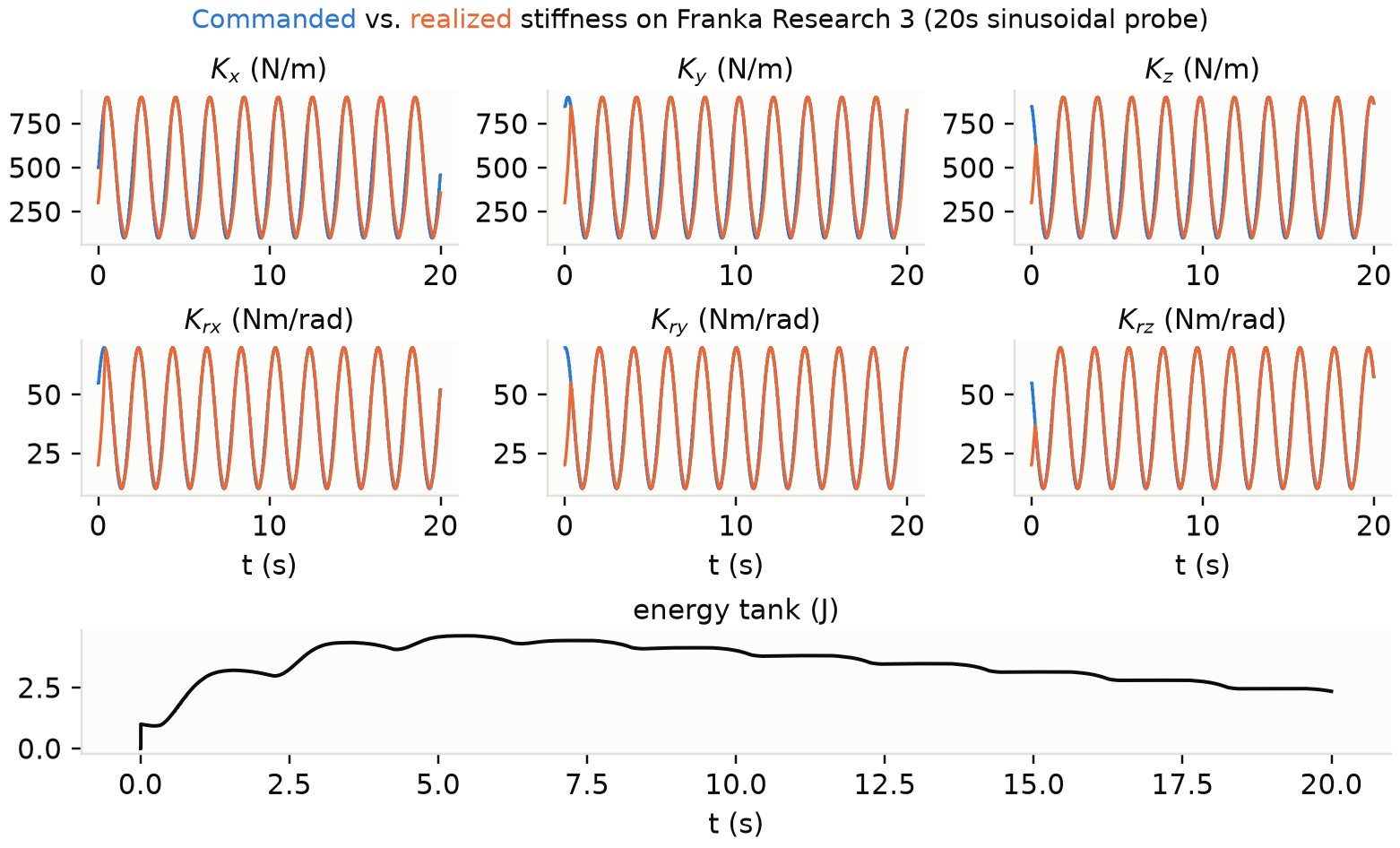}
    \caption{Commanded vs. realized stiffness under a 20s sinusoidal probe across all six axes, with the accompanying energy-tank trace, recorded on the physical follower arm.}
    \label{fig:controller_validation}
\end{figure}

Before evaluating policies we verify the controller responds to a commanded stiffness at all. Fig.~\ref{fig:controller_validation} shows commanded vs.\ applied stiffness under a $20$\,s sinusoidal probe on all six axes together with the energy-tank trace, on the physical follower arm.

Because all force metrics come from the same estimator on the same hardware, tool, payload and workspace, its bias is \emph{common-mode} across policies. The relative comparisons are valid, absolute values carry an offset, and we report all such quantities as estimated wrench.

\subsection{Instruction-conditioned compliance}
\label{sec:results-chain}
The paper's central claim is that compliance extracted from bilateral demonstrations is a \emph{learnable, instruction-conditioned action}. We test it as a three-link chain, each link measured separately on the manner contrast (\emph{firmly} vs.\ \emph{normally}) along the in-plane contact axis $f_x$ (Table~\ref{tab:chain}).

\begin{table}[t]
\centering
\caption{Instruction-conditioned compliance, measured end to end. Each link is a separate measurement on the same manner contrast; $d$ is Cohen's $d$ and $p$ is a two-sided Mann--Whitney $U$ test. Stiffness rows are geometric means and their $d$ is computed on $\log K_x$. The force row is in linear units. Units: $K_x$ in N/m, force in N.}
\label{tab:chain}
\begin{threeparttable}
\begin{tabular}{lccrr}
\toprule
Link & firmly & normally & $d$ & $p$ \\
\midrule
Demonstrated $K_x$\tnote{a} & 146.2 & 127.2 & 0.76 & 0.002 \\
Commanded $K_x$\tnote{b}    & 141.2 & 116.4 & 0.98 & 0.019 \\
Realized force\tnote{c}     &   9.12 &  6.40 & 0.89 & 0.023 \\
\bottomrule
\end{tabular}
\begin{tablenotes}\footnotesize
\item[a] Geometric mean of masked $K_x$: per episode, $\exp$ of the mean of
masked $\log K_x$; then pooled the same way across episodes. Primary
operator's demonstrations, $n=48$ firmly and $39$ normally episodes of the
$96$ collected (Sec.~\ref{sec:results-labels}).
\item[b] Same statistic on the policy's own $\log K_x$ output over executed
chunk steps ($n=12$ rollouts/manner, 3 seeds pooled).
\item[c] Per-rollout RMS estimated contact force ($n=12$ rollouts/manner, 3
seeds pooled).
\end{tablenotes}
\end{threeparttable}
\end{table}

All three links carry the instruction. The demonstrations encode the adverb in the in-plane stiffness ($d=0.76$); the policy reproduces that dependence in its own output ($d=0.98$); and the dependence survives all the way to the arm's realized contact force ($d=0.89$). The effect appears on the in-plane axes and not on the board normal ($f_z$ label $d=0.05$), which is what the physics predicts: against a rigid board, pressing harder principally moves the \emph{equilibrium penetration depth} $x_{\mathrm{eq}}$, whereas wiping firmly manifests as increased resistance to being deflected off the stroke path.

Two properties of the commanded stiffness are worth stating, as both are free of any rollout-budget confound. First, the policy commits to genuine anisotropy, that it is stiff along some directions and soft along others, rather than uniformly stiff: median commanded $\mathbf{K} = (126.2, 83.0, 209.1)$\,N/m in the contact frame, a per-timestep in-plane/normal ratio of $2.50$ (median; $6.63$ at the $95$th percentile), against only ${\approx}1.2\times$ in the pooled marginal label distribution. The policy is therefore not regressing to the label mean, a constant predictor could not produce this. Second, the effect is specific to compliance \emph{output}: Table~\ref{tab:m6} shows that no baseline modulates realized force with the instruction, including Force-In, which consumes force as an input, and Hybrid, which does move its commanded $\mathbf{K}$ with manner ($d=2.26$ on $f_y$) but whose realized force does not follow ($d=0.17$). Predicting compliance is not sufficient; the rest of the policy has to track it.

\begin{table}[t]
\centering
\caption{Does the manner word change what the arm actually does? Per-rollout
RMS estimated contact force (N), $n=12$ per manner per policy; $d$ is Cohen's
$d$ and $p$ a two-sided Mann--Whitney $U$ test, as in Table~\ref{tab:chain}.
Only the compliance-output policy responds.}
\label{tab:m6}
\begin{tabular}{lccrr}
\toprule
Policy & firmly & normally & $d$ & $p$ \\
\midrule
Fixed-K       & 10.69 & 11.20 & $-0.22$ & 0.977 \\
Fitted-K      &  6.78 &  6.23 &    0.24 & 0.214 \\
Force-In      & 13.23 & 12.58 &    0.10 & 0.436 \\
Hybrid        &  7.40 &  6.87 &    0.17 & 0.507 \\
\textbf{Ours} & \textbf{9.12} & \textbf{6.40} & \textbf{0.89} & \textbf{0.023} \\
\bottomrule
\end{tabular}
\end{table}

\subsection{Task performance}
Table~\ref{tab:e1-primary} reports in-distribution performance for all five
policies; Fig.~\ref{fig:qualitative_rollouts} shows representative rollouts of
our policy across the instruction grid.

\begin{table}[t]
\centering
\caption{In-distribution task performance and contact-force behavior
($n=24$ per policy, 3 seeds).}
\label{tab:e1-primary}
\footnotesize
\setlength{\tabcolsep}{3pt}
\begin{tabular}{lccccc}
\toprule
Policy & Success & Ink rem.\ (\%) & Peak (N) & RMS (N) & Prot. Stops \\
\midrule
Fixed-K       & 33.3\% & 32.6 $\pm$ 34.9 & 34.9 $\pm$ 14.0 & 10.9 $\pm$ 2.2 & 6/24 \\
Fitted-K      &  0.0\% &  1.3 $\pm$\phantom{0} 3.0 & 21.4 $\pm$\phantom{0} 7.2 &  6.5 $\pm$ 2.3 & 0/24 \\
Force-In      & 41.7\% & 37.3 $\pm$ 31.5 & 32.3 $\pm$\phantom{0} 9.0 & 12.9 $\pm$ 6.5 & 2/24 \\
Hybrid        & 20.8\% & 18.4 $\pm$ 28.8 & 19.6 $\pm$\phantom{0} 8.2 &  7.1 $\pm$ 3.1 & 0/24 \\
\textbf{Ours} & \textbf{50.0\%} & \textbf{40.8 $\pm$ 35.3} & 26.3 $\pm$\phantom{0} 9.1 &  7.8 $\pm$ 3.3 & 1/24 \\
\bottomrule
\end{tabular}
\end{table}

Contact-force behavior is more informative than success. Our policy attains the highest ink removal at a peak force $8.6$\,N lower and an RMS force $3.1$\,N lower than Fixed-K, and trips one protective stop in $24$ rollouts against Fixed-K's six. Hybrid is gentler still but removes less than half as much ink: the interesting axis is task progress \emph{at matched force}, and on that axis compliance output dominates both the fixed-stiffness and the force-input position policies.

\subsection{Is the force channel used?}
\label{sec:results-m7}
A policy can be given force and still ignore it. We freeze the force-history token at its first-observed value for the remainder of the episode and re-run the three policies that consume force; freezing is a no-op for Fixed-K and Fitted-K. Success falls for all three, but no drop is significant (Table~\ref{tab:m7}).
\begin{table}[t]
\centering
\caption{Causal force ablation. Force history frozen at its first-observed
value for the rest of the episode, vs.\ the live condition (Table~\ref{tab:e1-primary}).
Fixed-K and Fitted-K are excluded --- freezing is a no-op for both.}
\label{tab:m7}
\begin{tabular}{lccr}
\toprule
Policy & Live & Frozen & $\Delta$ \\
\midrule
Force-In & 41.7\% (10/24) & 35.7\% (5/14) & $-6.0$ pts \\
Hybrid   & 20.8\% (\phantom{0}5/24) & 16.7\% (2/12) & $-4.2$ pts \\
Ours     & 50.0\% (12/24) & 41.7\% (5/12) & $-8.3$ pts \\
\bottomrule
\end{tabular}
\end{table}
The result is inconclusive rather than negative. The three policies degrade almost identically in relative terms, so our policy's larger absolute drop follows from its higher baseline rather than from any greater reliance on force. The paper's central claim rests on Tables~\ref{tab:chain} and~\ref{tab:m6}, not on this ablation.

\subsection{Offline compliance-prediction benchmark}
\label{sec:results-benchmark}
A learned predictor beats a per-axis constant predictor on a held-out session on all three translational axes, for a $1.25\times$ aggregate improvement in $\log \mathbf{K}$ RMSE ($1.39\times$ on $f_x$, $1.03\times$ on $f_y$, $1.41\times$ on $f_z$). The rotational axes are excluded from the aggregate: their targets are box-saturated at the lower bound (Sec.~\ref{sec:limitations}), so their RMSE is small for reasons that carry no signal.

\subsection{Label validity on real hardware}
\label{sec:results-labels}
Applied to the primary operator's demonstrations, the identifiability mask retains 84.3--96.2\% of in-contact timesteps on the three translational axes, and 71.5--82.3\% on the rotational axes (Fig.~\ref{fig:mask_coverage}). The per-axis noise floor $\sigma_{f,i}$ that the mask's SNR condition consumes (Eq.~\eqref{eq:noise-floor}) is measured on held-out free-space sweep sessions and lies well below the labels' working range on every axis (Fig.~\ref{fig:noise-floor}). Extracted stiffness traces with phase annotation are shown in Fig.~\ref{fig:stiffness-traces}.

\subsection{Constant-stiffness fit (Fitted-K)}
\label{sec:results-oracle}
To place the label distribution in physically interpretable units we grid-search, per axis, the single constant stiffness minimizing validation RMSE. This is a property of the \emph{dataset}, not of a trained policy. The three translational constants ($141.1$, $149.5$ and $158.4$\,N/m for $f_x$, $f_y$, $f_z$) are unremarkable and lie well inside the controller's realizable $[50, 1500]$\,N/m range. All three rotational constants, however, land on the extraction's $5$\,N$\cdot$m/rad lower bound, for the reason given in Sec.~\ref{sec:limitations}. The rotational impedance present in these demonstrations is softer than the controller's minimum commandable rotational stiffness.

This is why Fitted-K must not be read as a tuned-impedance oracle. Deployed, its rotational constants leave the tool unable to hold orientation against the board, which accounts for its $0\%$ success rather than any property of constant impedance as such. The constant-impedance control in this study is Fixed-K, whose hand-set $(300,300,300,20,20,20)$ is well posed and which reaches $33.3\%$. We therefore do not claim that a well-tuned constant impedance is inadequate for this task; we claim, and Table~\ref{tab:m6} shows, that it cannot modulate force on instruction.

\section{DISCUSSION}
\label{sec:discussion}

Observing force and acting compliantly are different capabilities. The ForceVLA-style baseline sees the contact wrench at every timestep, yet produces rigid position targets, leaving one way to change contact against an unyielding board is to drive the equilibrium deeper. It records the highest RMS force of any policy, $12.9$\,N, without removing more ink than ours. Fixed-K, with no force channel, meets the board with a constant that cannot yield and trips six protective stops in $24$ rollouts, where in our case modulating $\mathbf{K}$ holds contact at $8.6$\,N lower peak force with one stop. Predicting compliance is not by itself enough,although Hybrid changes its commanded stiffness with the instruction more strongly than we do, but decodes it through a regression head separate from its pose, and the two never co-vary at execution, so its realized force does not follow. Decoding both through one shared target is what closes that gap, and our policy commits to genuine anisotropy as a result, a median in-plane to normal ratio of $2.50$ against roughly $1.2$ in the pooled labels. Task success separates the policies far less cleanly, $50.0\%$ against Force-In's $41.7\%$ is two rollouts in $24$, and a well-posed Fixed-K still completes the task a third of the time. What survives the small sample is categorical rather than quantitative. Ours is the only policy whose realized contact force follows the language instruction (Table~\ref{tab:m6}), the policy learned that a firm wipe resists deflection off the stroke, not that it presses harder. 

\section{Limitations}
\label{sec:limitations}

Sensorless wrench estimation is configuration-dependent, so we pin a fixed nullspace posture and constrain the working volume during calibration and collection. A wrist F/T sensor would remove that constraint at the cost of bulk, wiring and expense. The rotational channels are weakly identified as our wiping task barely excites orientation, and the rotational impedance the operators used is softer than our controller's minimum commandable stiffness, so $\tau_x$, $\tau_y$ and $\tau_z$ saturate at the lower bound and should be read as ``at least this soft'' rather than as calibrated values. That reflects this task and this controller's realizable range, not the identifiability argument, which holds per axis wherever the axis is excited. The evidence is also narrow (one task, one platform) and the rollout budget is small enough that the force-channel ablation (Sec.~\ref{sec:results-m7}) is underpowered by more than an order of magnitude, leaving the causal role of the force input open.

\section{CONCLUSION}
\label{sec:conclusion}

Pose-only teleoperation cannot supply compliance supervision, because realized pose and measured force leave the operator's intended equilibrium and their stiffness confounded. Four-channel bilateral teleoperation breaks the tie as the leader arm measures the intended equilibrium directly, which makes per-axis stiffness identifiable from demonstrations that are already being collected, with no force-torque hardware and no additional annotation. The resulting labels are learnable enough to drive a VLA policy whose contact force responds to a language instruction, which no baseline here achieves. Extending the extraction to dynamic, impulsive contact, and to cheaper embodiments that lack joint-torque sensing, is left to future work.

\FloatBarrier
\bibliographystyle{IEEEtran}
\bibliography{references}

\end{document}